\documentclass[11pt]{article}

\usepackage[final]{acl}

\usepackage{times}
\usepackage{latexsym}

\usepackage{dblfloatfix}
\usepackage[T1]{fontenc}

\usepackage[utf8]{inputenc}

\usepackage{microtype}

\usepackage{inconsolata}

\usepackage{graphicx,hyperref}

\usepackage{amssymb}
\usepackage{booktabs}
\newcommand{\name}{\textsc{KAIO}}

\newcommand{\titleicon}[2][1.2em]{%
  \raisebox{-.2\height}{\includegraphics[height=#1]{#2}}}

\title{\texorpdfstring{%
  \titleicon[2.4em]{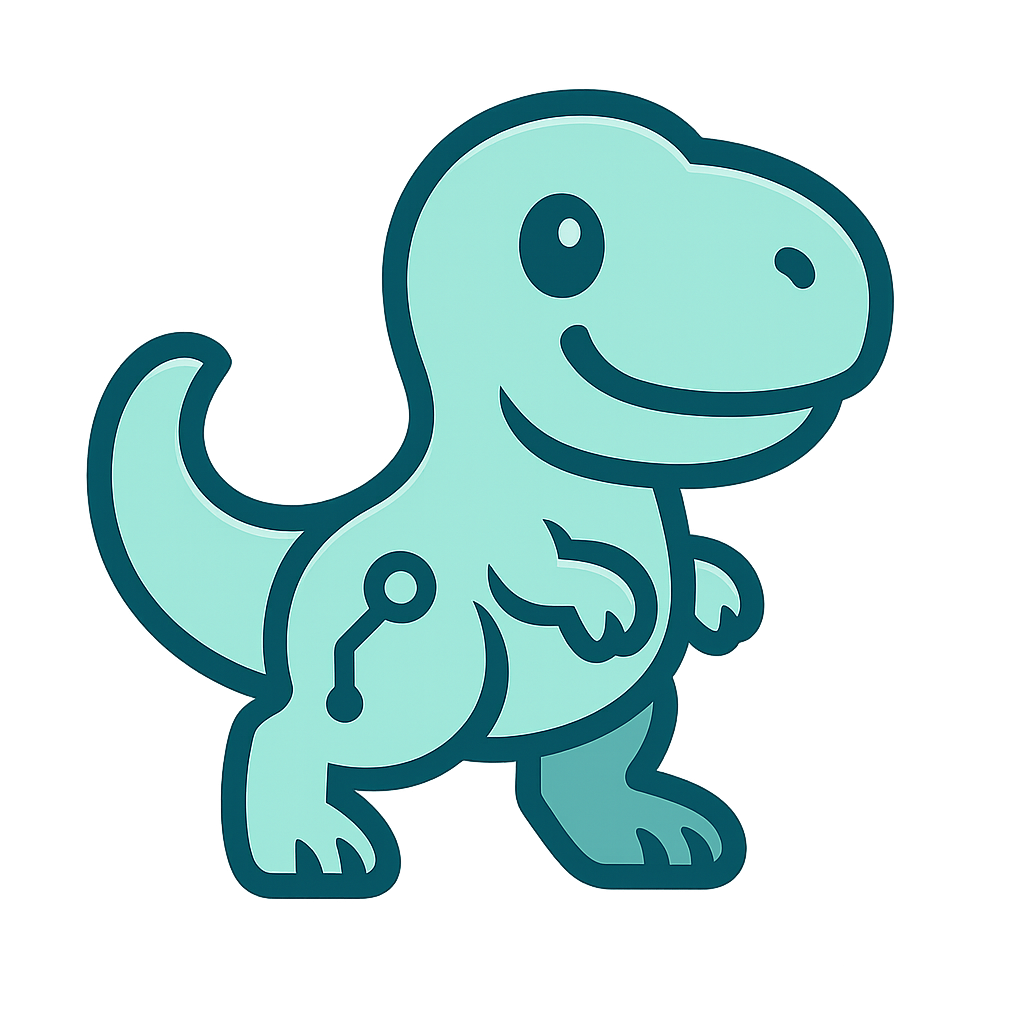}\ \name{}: A Collection of \textit{More} Challenging Korean Questions%
}{\name{}: A Collection of More Challenging Korean Questions}}

\author{
  Nahyun Lee$^{1,4}$ \\
  Chung-Ang University$^{1}$ \\
  MODULABS$^{4}$ 
  \And
  Guijin Son$^{2,4}$ \\
  OneLineAI$^{2}$ \\
  MODULABS$^{4}$
  \And
  Hyunwoo Ko$^{2,4}$ \\
  OneLineAI$^{2}$ \\
  MODULABS$^{4}$ \\
  \And
  Kyubeen Han$^{3,4}$ \\
  Konkuk University$^{3}$ \\
  MODULABS$^{4}$ \\
}

\author{
Nahyun Lee{\textsuperscript{1,4}} \quad 
Guijin Son{\textsuperscript{2,4}} \quad 
Hyunwoo Ko{\textsuperscript{2,4}} \quad
Kyubeen Han{\textsuperscript{3,4}}
\\[0.5em]  
Chung-Ang University{\textsuperscript{1}} \quad 
OneLineAI{\textsuperscript{2}} \quad 
Konkuk University{\textsuperscript{3}} \quad 
MODULABS{\textsuperscript{4}}
}

\begin{document}
\maketitle


\begin{abstract}
With the advancement of mid/post-training techniques, LLMs are pushing their boundaries at an accelerated pace. Legacy benchmarks saturate quickly (e.g., broad suites like MMLU over years, newer ones like GPQA-D even faster), which makes frontier progress hard to track. The problem is \emph{especially acute in Korean}: widely used benchmarks are fewer, often translated or narrow in scope, and updated more slowly, so saturation and contamination arrive sooner. Accordingly, at this moment \emph{there is no Korean benchmark} capable of evaluating and ranking frontier models. To bridge this gap, we introduce \textbf{KAIO}, a Korean, math-centric benchmark that stresses long-chain reasoning. Unlike recent Korean suites that are at or near saturation KAIO remains far from saturated: the best-performing model, GPT-5, attains 62.8 followed by Gemini-2.5-Pro (52.3). Open models such as Qwen3-235B and DeepSeek-R1 cluster falls below 30 demonstrating substantial headroom, enabling robust tracking of frontier progress in Korean. To reduce contamination, KAIO will remain private and be served via a held-out evaluator until the best publicly known model reaches at least \textbf{80}\% accuracy, after which we will release the set and iterate to a harder version.
\end{abstract}

\section{Introduction}

Traditionally, major LLM revisions were slow and costly: advancing a model typically required collecting and cleaning massive corpora~\citep{bai2023qwen, gao2020pile} followed by billion-dollar–scale pre-training~\citep{openai_gpt5_system_card_2025}. Recent mid- and post-training methods (e.g., continued pre-training, SFT, preference optimization, RL/distillation)~\citep{guo2025deepseek} now enable frequent, lower-cost updates. With faster updates, however, benchmarks saturate more often. For example, \textit{MMLU} (released 2020/09)~\citep{hendrycks2020measuring} took \(\approx31\) months to see the first \(>80\%\) result (GPT-4, 2023/03), whereas \textit{GPQA-Diamond} (released 2023/11)~\citep{rein2024gpqa} crossed \(80\%\) in \(\approx15\) months (2025/02) despite being substantially harder. 

\begin{figure}[t]
\includegraphics[width=1\linewidth]{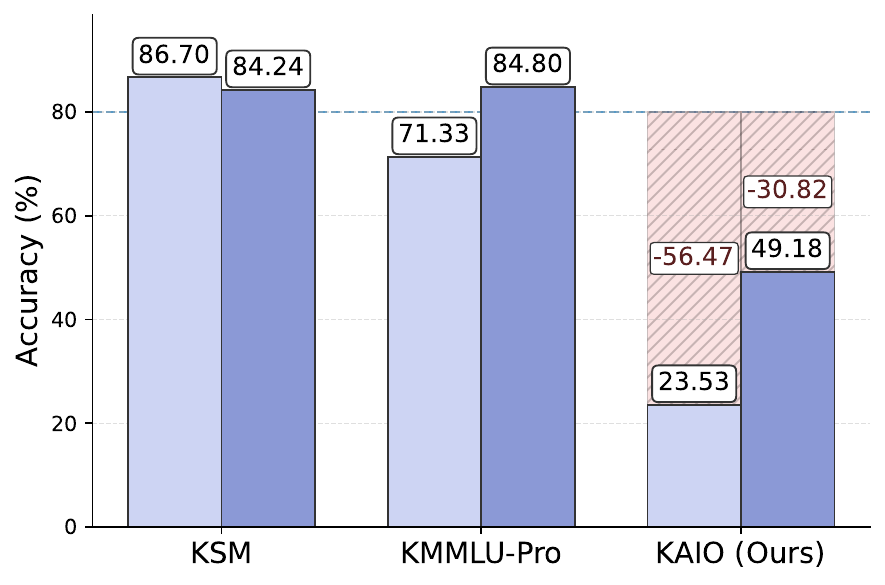}
\centering
\caption{\footnotesize Accuracy of DeepSeek-R1 (light purple) and Gemini-2.5-Pro (dark) on KSM, KMMLU-Pro, and KAIO.
}
\label{fig:benchmark_saturation}
\end{figure}

The problem is sharper in Korean: benchmark releases are scarce, and modern multilingual models now perform in Korean nearly as well as in English, rapidly saturating existing suites. As Figure~\ref{fig:benchmark_saturation} illustrates, the math benchmark \textsc{KSM} (Jan 2025)~\citep{ko2025understand} and the general-knowledge \textsc{KMMLU-Pro} (May 2025)~\citep{hong2025kmmlu} already have models exceeding 80\%, for \textsc{KMMLU-Pro}, within roughly three months of release. This pace highlights the need for harder, contamination-resistant Korean benchmarks.

To this end, we introduce \name{}, a Korean mathematics benchmark designed to stress top-tier LLMs. We manually curated \textbf{331} candidate problems from publicly available, high-difficulty sources (e.g., contest and olympiad preparation sites). We then applied multi-stage adversarial filtering, de-duplication, ambiguity and leakage checks, and model-in-the-loop screening—retaining only items that consistently defeat strong frontier models. The final set contains \textbf{44} problems selected for maximal difficulty.

Our evaluation shows that, the best-performing model, GPT-5, attains 62.8, followed by Gemini-2.5-Pro (52.3) and open models clustering below 30. On efficiency, GPT-5 sits on the accuracy–cost frontier at ~7.8k tokens per query. Finally, we observe that output length generally does not predict correctness (Welch’s $t$-test, $\alpha{=}0.05$). To prevent contamination, KAIO will remain private and be served via a held-out evaluator until the best publicly known model reaches at least \textbf{80}\% accuracy, after which we will release the set and iterate to a harder version.

\section{Preliminary: Saturation of Korean Benchmarks}


\begin{table}[t]
\centering
\fontsize{8.5}{9.5}\selectfont
\begin{tabular}{lcc}
\toprule
\textbf{Benchmark} & \textbf{Release date} & \textbf{$\ge$80\% reached} \\
\midrule
Ko-HellaSwag       & Nov 2023               & \checkmark 
\\
Ko-TruthfulQA      & Feb 2024               & \checkmark  
\\
KMMLU              & Feb 2024               & \checkmark  
\\
KSM                & Jan 2025               & \checkmark  
\\
KMMLU-Pro          & May 2025               & \checkmark  
\\
\bottomrule
\end{tabular}
\caption{\footnotesize Saturation status of major Korean-language benchmarks. Each row lists the benchmark, its first public release date, and whether any language model has achieved 80\% accuracy}
\label{tab:korean_benchmarks_80}
\end{table}


In recent years, LLMs have rapidly ascended toward high-level performance, demonstrating impressive proficiency not only in English but also in Korean.
When benchmark performance transcends the 80\% threshold, its interpretive value often diminishes due to inherent limitations in dataset curation and evaluation methodology~\cite{kiela2021dynabench, gema2024we}. 
In this context, distinctions such as 85\% vs. 90\% are likely influenced more by dataset artifacts or annotation noise than by genuine differences in model capability.
At these elevated performance levels, the remaining errors frequently stem from ambiguous problem statements, annotation inconsistencies, or subjective interpretation differences rather than genuine model capabilities~\citep{northcutt2021pervasive, bowman-dahl-2021-will, kovatchev-lease-2024-benchmark}. 

Earlier Korean benchmarks~\citep{park2024open} such as \textsc{Ko-HellaSwag} (Nov. 2023) and the factuality set \textsc{Ko-TruthfulQA} (Feb. 2024)—saturated rapidly, with leading systems exceeding 80\% accuracy within $\sim$25 weeks of release. More recent datasets have converged even faster: the mathematics benchmark \textsc{KSM} (Jan. 2025)~~\citep{ko2025understand} reached saturation in about six months, while the general-knowledge suite \textsc{KMMLU-Pro} (May 2025)~\citep{hong2025kmmlu} crossed the 80\% threshold in roughly three months.

To address these fundamental limitations, a reorientation of benchmark design philosophy is essential. Instead of pursuing comprehensive coverage through large-scale datasets, rigorous evaluation of frontier models necessitates meticulously curated problems that maintain discriminative power even at high performance levels. This methodology emphasizes \emph{depth over breadth}, concentrating on issues that persistently challenge state-of-the-art systems.



\section{\name{}}
\subsection{Dataset Collection}

We systematically collected 331 candidate problems from mathematics contests and olympiad preparation websites. Problems were digitized via a YOLO-based OCR pipeline and subsequently validated by five human reviewers using a structured annotation interface (Figure~\ref{fig:annotation_tool}), ensuring accurate mathematical notation and eliminating LaTeX artifacts. We then applied multi-stage difficulty screening: each candidate was evaluated with fixed snapshots of earlier state-of-the-art models (e.g., GPT-4.1, Gemini-2.0-Pro~\citep{comanici2025gemini}, Qwen2.5-72B~\citep{yang2024qwen2}, Llama-3.1-70B~\citep{grattafiori2024llama}). Using prior SOTA systems, rather than the contemporary models under evaluation, reduces the risk of adversarial tailoring and avoids unintended penalties to specific model families. We retained only problems that remained consistently challenging across diverse architectures. This process yields a final set of 44 problems combining high mathematical complexity with precisely rendered content. The size is comparable to recent math benchmarks, such as AIME 2024/2025, which offer two 15-question rounds (A and B) per year for a total of 30 questions.

\subsection{Dataset Description}

\begin{table}[t]
\centering
\fontsize{9}{10.5}\selectfont
\begin{tabular}{@{}p{\columnwidth}@{}}
\toprule
\textbf{Linear Algebra} \\
\midrule
For a positive integer $n$, let $A$ be an $n \times n$ real matrix with $n$ distinct positive eigenvalues. Find the number of real matrices $B$ satisfying $B^{2016} = A$. \\
\emph{Answer:} $2^n$ \\
\midrule
\textbf{Geometry \& Analysis} \\
\midrule
Compute the area of the quadrilateral with vertices $(0,0,0)$, $(1,2,3)$, $(3,1,2)$, $(7,4,7)$. \\
\emph{Answer:} $\frac{15\sqrt{3}}{2}$ \\
\bottomrule
\end{tabular}%
\caption{\footnotesize \textbf{Example question–answer blocks.} Each block lists a category, a representative question, and its answer.}
\label{tab:dataset_examples}
\end{table}

\begin{figure*}[!t]
\centerline{\includegraphics[width=\textwidth]{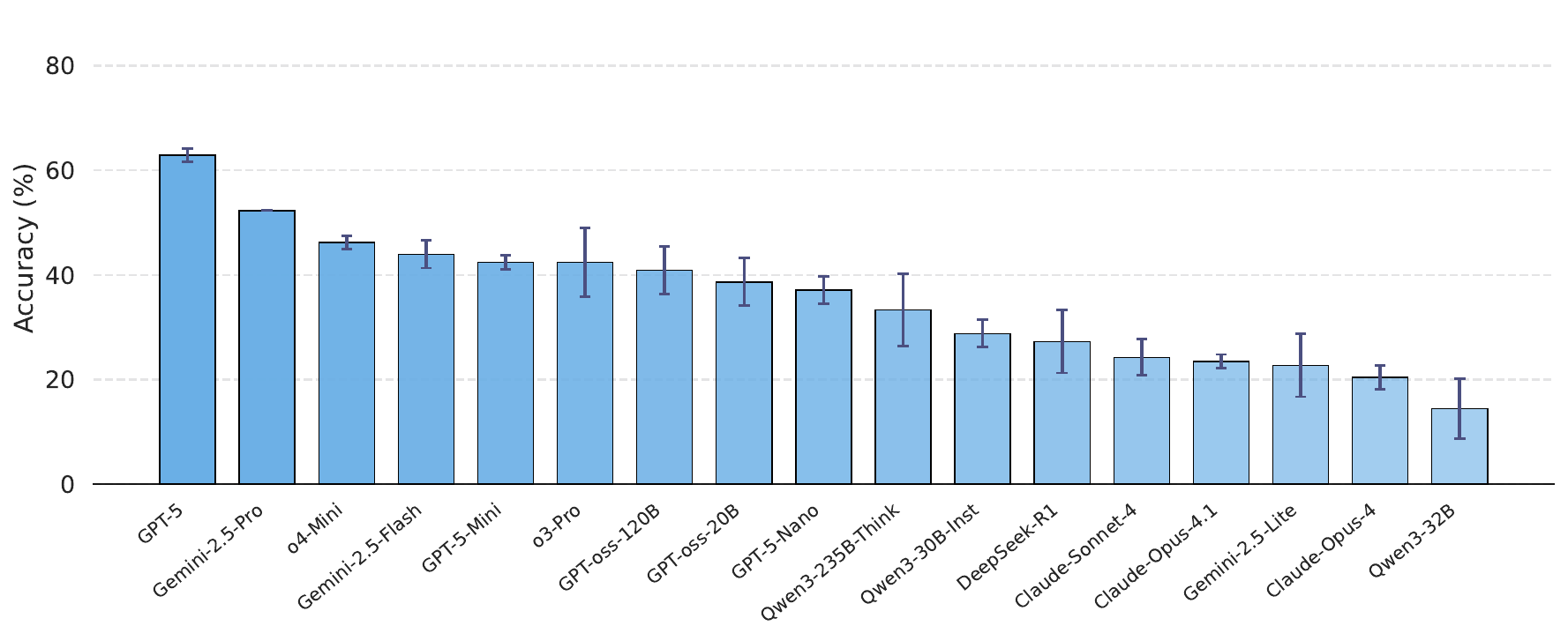}}
\caption{\textbf{KAIO benchmark results showing model accuracy with standard deviation across three runs.} GPT-5 leads at 62.9\%, with clear performance tiers emerging across model families. Error bars represent standard deviation from three independent evaluations.}
\label{fig:model_performance_accuracy}
\end{figure*}

Table~\ref{tab:dataset_examples} shows examples of questions included in KAIO. 
The KAIO dataset covers a wide range of problem types, including linear algebra, analysis, probability and statistics, optimization, and logical reasoning. 
In terms of input size, the questions average 216 tokens, spanning from short problems of 38 tokens to longer ones of up to 1,950 tokens. 
This diversity in both domain and length makes the dataset well-suited for evaluating models across a broad spectrum of mathematical and scientific reasoning skills, from short factual derivations to extended multi-step solutions.

\section{Main Results}

\subsection{Evaluation Setup}
We conduct a comprehensive evaluation of each model across the complete KAIO dataset to assess both performance capabilities and response consistency.
We evaluated 17 models spanning major model families, including top-performing systems such as \textit{GPT-5}~\cite{openai_gpt5_system_card_2025}, \textit{Claude-Opus-4.1}~\citep{anthropic_claude_opus}, and \textit{Qwen3-235B-Think}~\citep{yang2025qwen3}, alongside broader families such as the OpenAI o series~\citep{openai_o3_o4mini_2025}, \textit{Gemini-2.5}~\citep{comanici2025gemini} and DeepSeek-R1~\citep{guo2025deepseek}. 
For each model, we conducted three independent runs to account for stochasticity in generation\footnote{We use temperature 0.9 and top\_p 0.95.}. All models were evaluated under a standardized protocol with identical prompt formatting and decoding settings. Accuracy is reported with corresponding mean and standard deviation values. Finally, we use \textit{GPT-4.1} strictly as an answer extractor and normalizer: it converts each model’s output to a canonical form and compares it with the gold label. It does not inspect or score the solution process or reasoning traces,its role is purely mechanical matching.
Additionally, we record total token consumption encompassing both input prompts and generated completions as a proxy for computational cost and reasoning verbosity.

\begin{figure}[t]
\includegraphics[width=1\linewidth]{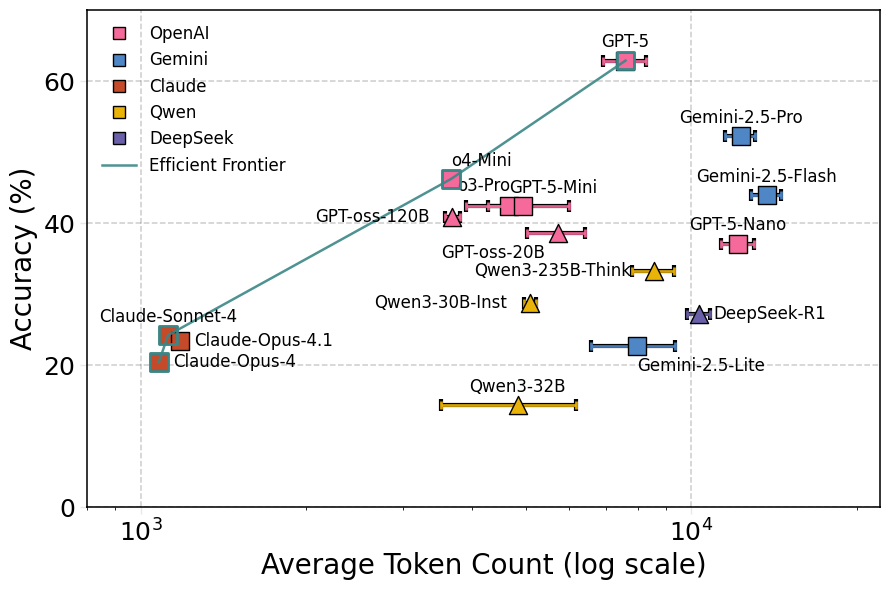}
\centering
\caption{\footnotesize \textbf{Scatter plot of model accuracy (y-axis) versus average token usage (x-axis, log scale).} Each point represents a model, with horizontal bars denoting the standard deviation of token counts. Colors indicate model families, while marker shapes distinguish open-source models shown as triangles from closed-source models shown as squares. The dashed line highlights the efficient frontier, connecting models that achieve the best trade-off between accuracy and token efficiency.}
\label{fig:accuracy_and_token_counts}
\end{figure}

\subsection{Model Accuracy Comparison}
Figure \ref{fig:model_performance_accuracy} reports model accuracy on KAIO along with standard deviations over three runs. GPT-5 achieves 62.8\% accuracy, showing substantial separation from competing architectures. Gemini-2.5-Pro and o4-Mini form the subsequent high-performing cluster, reaching 52.3\% and 46.2\% accuracy, respectively. In the lower performance regime, disparities among models become more pronounced. Qwen3-235B, Qwen3-30B, and DeepSeek-R1 converge within the 27–33\% accuracy range, whereas the Claude family underperforms relative to peer architectures, attaining only 20–24\% accuracy. Notably, Qwen3-32B falls below 15\% trailing behind smaller models (Qwen3-30B) of the same family.

\subsection{Token Efficiency and Model Accuracy}
Figure \ref{fig:accuracy_and_token_counts} illustrates model accuracy as a function of average tokens consumed per query on a logarithmic scale, with horizontal error bars denoting variance across experimental iterations. The solid curve traces the efficiency frontier, encompassing models that achieve superior trade-offs between accuracy and computational cost. GPT-5 stands at the frontier’s apex of this frontier, attaining the highest accuracy while maintaining reasonable token consumption at approximately 7.8k tokens per query. The subsequent position along the efficiency continuum is occupied by o4-Mini, which demonstrates strong performance-cost characteristics by achieving 46.2\% accuracy with markedly economical token usage of approximately 3.9k tokens per query.

\begin{figure}[t]
\includegraphics[width=1\linewidth]{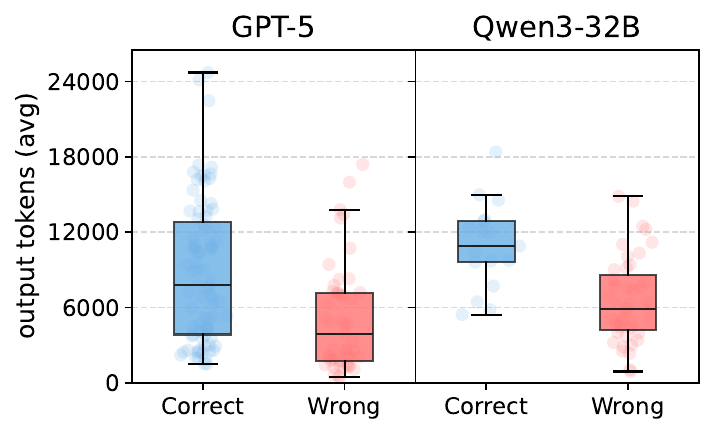}
\centering
\caption{\footnotesize \textbf{Output tokens vs. correctness (GPT-5, Qwen3-32B).} Boxplots compare output-token counts for correct vs.\ incorrect answers. While most models show no significant difference by Welch’s t-test, GPT-5 and Qwen3-32B generate significantly longer outputs when the answer is correct.}

\label{fig:benchmark_token}
\end{figure}

\subsection{Output token count and Correctness} 
For each model, we tested whether the number of output tokens differs between correct and incorrect responses using a two-sided Welch's t-test ($\alpha=0.05$). Across the suite of models, we found no statistically detectable difference in most cases, indicating that generating more or fewer tokens does not consistently correlate with correctness for the majority of models.
However, two models stand out as exceptions (Figure \ref{fig:token_vs_correct_per_model}). GPT-5 produces more tokens when giving correct answers compared to incorrect ones, with correct responses averaging 8,807 tokens versus 5,532 for incorrect ones,  suggesting a moderate relationship between response length and accuracy for this model ($p=0.0012$, $Cohen’s~d=0.59$). By contrast, Qwen3-32B showed a stronger pattern, with correct answers averaging 10,931 tokens compared to 6,570 for incorrect responses ($p=0.0001$, $d=1.27$), indicating a clearer relationship. These are contrary to recent works~\citep{chen2024not} that models tend to overthink when underconfident. Detailed plots and statistical results are provided in Figure~\ref{fig:token_vs_correct_per_model}.

\section{Conclusion}
In this work, we introduced \textsc{KAIO}, a high-difficulty Korean benchmark designed to rigorously assess the reasoning capabilities of advanced LLMs. 
Unlike existing Korean benchmarks that quickly reach saturation, KAIO provides a carefully curated set of mathematical reasoning problems, emphasizing depth over breadth to preserve discriminative power. Our evaluation of 17 models reveal that KAIO presents a challenge to existing models with GPT-5 scoring 62.8\% and most open models falling below 30\%. KAIO will be kept private and updated whenever the leading models surpass the 80\% threshold, ensuring its continued relevance.  Ultimately, KAIO is intended as a foundation for advancing reliable reasoning in Korean and as a blueprint for benchmark development in other low-resource languages.



\bibliography{custom}

\appendix

\section{Additional Figures}
\label{sec:appendix}

\begin{figure*}[!t]
\centerline{\includegraphics[width=0.8\textwidth]{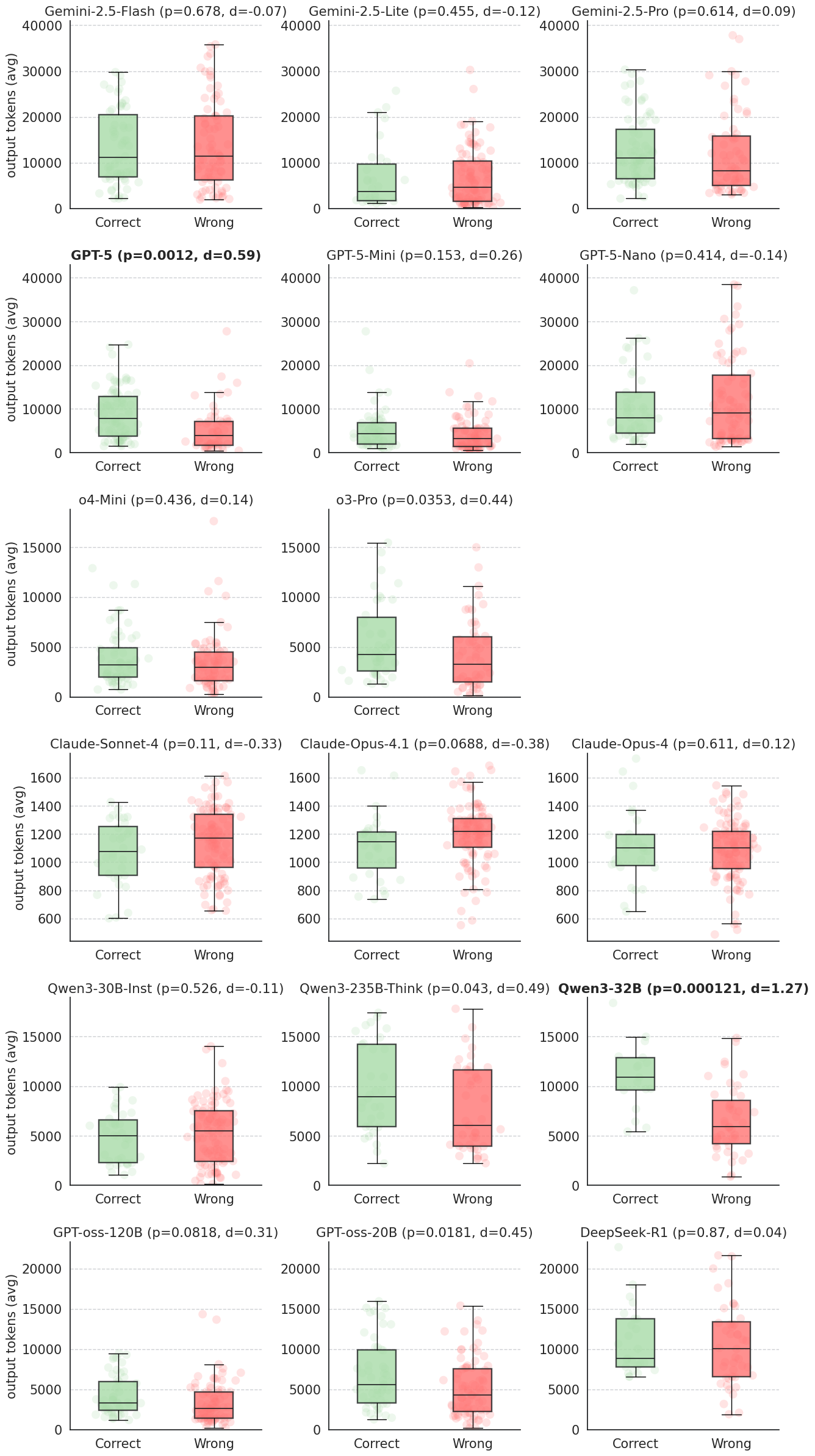}}
\caption{\textbf{Output tokens vs. correctness for each model family.} Boxplots compare output-token counts between correct and incorrect answers.}
\label{fig:token_vs_correct_per_model}
\end{figure*}

\begin{figure*}[!t]
\centerline{\includegraphics[width=0.8\textwidth]{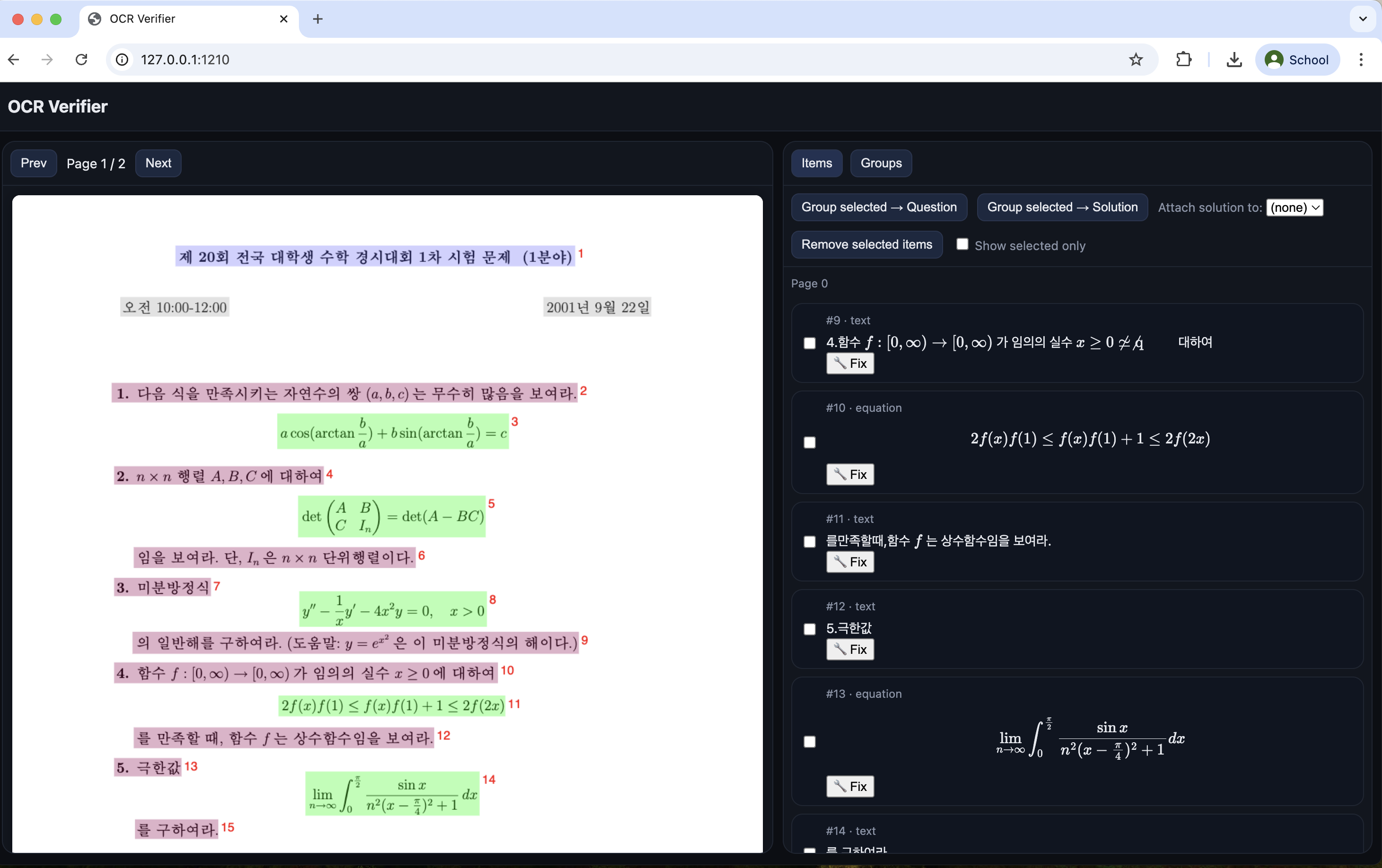}}
\caption{\textbf{Annotation tool used for OCR verification.}}
\label{fig:annotation_tool}
\end{figure*}

\end{document}